\documentclass[10pt,conference]{IEEEtran}
\IEEEoverridecommandlockouts

\usepackage[frozencache,cachedir=.]{minted}
\usepackage{cite}
\usepackage{amsmath,amssymb,amsfonts}
\usepackage{algorithmic}
\usepackage{graphicx}
\usepackage{textcomp}
\usepackage{xcolor}  
\usepackage{hyperref}
\usepackage{multirow}
\usepackage{multicol}
\usepackage{caption}
\usepackage{setspace}
\usepackage{adjustbox}
\usepackage{dblfloatfix}
\usepackage{svg}
\usepackage{booktabs}
\usepackage{listings}
\usepackage{color}
\usepackage{enumerate}
\usepackage{subfigure}
\usepackage{rotating}
\usepackage{comment}
\usepackage{tabularx}
\usepackage{enumitem}
\usepackage{soul}
\usepackage{tikz}
\usetikzlibrary{tikzmark}
\usetikzlibrary{calc} 
\usetikzlibrary{arrows}
\usetikzlibrary{shapes}

\newcommand{\ie}{\textit{i.e.},\ }

\newcommand{\etal}{\textit{et al.} }
\newcommand{\etc}{{\em etc.}}

\usepackage{booktabs}

\definecolor{francBlue}{RGB}{64,76,87}

\usepackage[most]{tcolorbox}
\usepackage[parfill]{parskip}

\tcbset{
  parskip/.style={
    before={\par\pagebreak[0]\parindent=0pt},
    after={\parfillskip=0pt\par},
  },
}
\newtcolorbox{resultbox}[1][]{%
    colback=black!3,
    colframe=black!3,
    notitle,
    sharp corners,
    borderline west={2pt}{0pt}{gray!80!black},
    enhanced,
    breakable,
    boxsep=0pt,
    left=4pt,right=2pt,top=2pt,bottom=2pt,
    }
\lstnewenvironment{mylisting}{%
    \lstset{%
        basicstyle=\small\ttfamily,
        frame=single,
        numbers=left,
        numberstyle=\tiny,
    }%
}{}

\definecolor{codebg}{rgb}{0.99,0.99,0.99}
\definecolor{hiliteColor}{rgb}{1,0.92549019607,0.6}
\definecolor{tainted}{rgb}{0,1,1}

\newmintinline{python}{fontsize=\normalsize}

\newminted[PythonSourceCode]{python}{
    labelposition=topline,
    frame=single,
    numbersep=2pt,
    fontsize=\tiny,
    xleftmargin=5pt,
    framerule=0.5pt,
    linenos
}

\newminted[JavaSourceCode]{java}{
    labelposition=topline,
    frame=single,
    numbersep=2pt,
    fontsize=\tiny,
    xleftmargin=5pt,
    framerule=0.5pt,
    linenos,
    escapeinside=||
}
\newminted[PlainTextCode]{plaintext}{
    labelposition=topline,
    frame=single,
    numbersep=2pt,
    fontsize=\tiny,
    xleftmargin=5pt,
    framerule=0.5pt,
    linenos,
    escapeinside=||
}
\definecolor{magnolia}{rgb}{0.97, 0.96, 1.0}
\definecolor{shadecolor}{rgb}{0.97, 0.96, 1.0}

\newmintinline[codePython]{python}{fontsize=\small}
\newmintinline[codeJava]{java}{fontsize=\small}
\newmintinline[codeHtml]{html}{fontsize=\small}
\newmintinline[snippetJava]{java}{fontsize=\small,bgcolor=magnolia}
\newmintinline[snippetPython]{python}{fontsize=\small,bgcolor=magnolia}
\newmintinline[snippetPerl]{perl}{fontsize=\small,bgcolor=magnolia}
\newmintinline[snippetHtml]{html}{fontsize=\small,bgcolor=magnolia}
\newcommand{\codeTag}[1]{\colorbox{magnolia}{\textbf{\texttt{<\textcolor{red}{#1}>}}}}

\definecolor{dkgreen}{rgb}{0,0.6,0}
\definecolor{gray}{rgb}{0.5,0.5,0.5}
\definecolor{mauve}{rgb}{0.58,0,0.82}

\usepackage[textwidth=1.8cm]{todonotes}          
\usepackage{marginnote}         
\usepackage{setspace}           
\newcounter{mycomment}
\newcommand{\mycomment}[2][]{%
\refstepcounter{mycomment}%
{%
\setstretch{0.7}
\todo[color={red!100!green!33},size=\tiny]{%
\textbf{#1\#\themycomment}~#2}%
}}

\newcommand{\jcss}[1]{\mycomment[JCSS]{#1}}
\newcommand{\think}

\newcommand{\negative}[1]{\textcolor{red}{$_{-#1}$}}
\newcommand{\positive}[1]{\textcolor{green}{$_{+#1}$}}

\def\BibTeX{{\rm B\kern-.05em{\sc i\kern-.025em b}\kern-.08em
    T\kern-.1667em\lower.7ex\hbox{E}\kern-.125emX}}

\begin{document}


\title{Enhancing Automated Program Repair through Fine-tuning and Prompt Engineering}


\author{
    \IEEEauthorblockN{Rishov Paul\IEEEauthorrefmark{1}, Md. Mohib Hossain\IEEEauthorrefmark{1},  Mohammed Latif Siddiq\IEEEauthorrefmark{2}\\ Masum Hasan\IEEEauthorrefmark{3}, Anindya Iqbal\IEEEauthorrefmark{1}, and  Joanna C. S. Santos\IEEEauthorrefmark{2}}
    \IEEEauthorblockA{\IEEEauthorrefmark{1}Department of Computer Science and Engineering, BUET, Dhaka, Bangladesh}
    \IEEEauthorblockA{\IEEEauthorrefmark{2}Department of Computer Science and Engineering, University of Notre Dame, USA}
    \IEEEauthorblockA{\IEEEauthorrefmark{3}Department of Computer Science, University of Rochester, USA}
    \\\{rishov.paul, mdmohib.hossain\}@iqvia.com, msiddiq3@nd.edu\\m.hasan@rochester.edu, anindya@cse.buet.ac.bd, and joannacss@nd.edu
}


\maketitle

\thispagestyle{plain}
\pagestyle{plain}

\begin{abstract}
Sequence-to-sequence models have been used to transform erroneous programs into correct ones when trained with a large enough dataset. Some recent studies also demonstrated strong empirical evidence that code review could improve the program repair further. Large language models, trained with Natural Language (NL) and Programming Language (PL), can contain inherent knowledge of both.
In this study, we investigate if this inherent knowledge of PL and NL can be utilized to improve automated program repair.
We applied PLBART and CodeT5, two state-of-the-art language models that are pre-trained with both PL and NL, on two such natural language-based program repair datasets and found that the pre-trained language models fine-tuned with datasets containing both code review and subsequent code changes notably outperformed each of the previous models. With the advent of code generative models like Codex and GPT-3.5-Turbo, we also performed zero-shot and few-shots learning-based prompt engineering to assess their performance on these datasets. However, the practical application of using LLMs in the context of automated program repair is still a long way off based on our manual analysis of the generated repaired codes by the learning models. 
\end{abstract}

\begin{IEEEkeywords}
automated program repair, pre-trained transformer model, code review, prompt engineering, GPT3
\end{IEEEkeywords}

\section{Introduction}

Code review is the process of analyzing source code that has been written by a collaborator in order to determine whether or not it is of sufficient quality to be merged into the main source code repository~\cite{bird2013expectations}. Code review provides several advantages, including improving the overall quality of the code and decreasing the likelihood of introducing errors into the system~\cite{Baum2016industry,mcintosh2016empirical}.

The defects identified by code reviewers, testers, or static analysis tools need to be fixed within a short deadline before the release of the software. However, repairing defects in a program is time-consuming and expensive. In fact, this process accounts for nearly half of the total cost and time of software development~\cite{reversibleDebugging}. Hence, automation of code repair can be highly beneficial for the software development sector.

Traditional automatic program repair approaches fix a program using test suites~\cite{auto_repair_buggy_if_missing_precondition, program_repair_semantic_analysis, human_written_patch, condition_synthesis}. However, it still takes extra work to construct these test suites. Alternative approaches, such as static analysis-based~\cite{semfix, SequenceR} and learning-based automated code repair techniques~\cite{getafix, deepfix, lutellier2020coconut}, have yet to achieve acceptable results. This has motivated researchers to develop solutions using code review suggestions to achieve a better quality of bug fix suggestions~\cite{ R4R, tufano21} . They have established that when a defective (``buggy'') code is given to repair, there is a performance boost if review comments are given alongside it. Since code review is a common practice\cite{McIntosh_code_review}, using them requires no additional resources. 
Despite considerable and promising improvement, the learning-based models presented in~\cite{tufano21, R4R} could not achieve sufficient ability to be used in industry-level code repair. For example, the accuracy of baseline models for the corresponding datasets is around 12\% to 20\%. 
Therefore, this research direction needs much further exploration to advance the state-of-the-art with the current techniques.

Recent works have successfully used  transformer-based~\cite{transformers} pre-trained models for different relevant software engineering tasks such as code summarization, code search, code documentation, code refinement \etc~\cite{PLBART, CodeT5, guo2021graphcodebert, feng-etal-2020-codebert}.
These models are trained on large corpora to acquire universal language representations. They may then be used for downstream NLP (Natural Language Processing) tasks without having to train new models from scratch, for which, nowadays, the transformer has become now the standard pre-trained model architecture. 
Hence, it is important to study whether these models can improve the results of program repair by effectively utilizing code review along with the associated code context.

Additionally, recent progress on large language models (LLMs)~\cite{LLM}, like GPT-3.5, has demonstrated excellence in producing code from well-formulated prompts. With the increasing popularity of LLMs, prior works have investigated the correctness of the generated code~\cite{dakhel2022github}, their quality (in terms of code smells)~\cite{siddiq2022empirical}, security~\cite{pearce2022asleep} as well as whether it can be used for API learning tasks~\cite{apiicpc22}, code complexity prediction~\cite{siddiq23zero}. Since they have exhibited strong zero-shot~\cite{zero_shot} and few-shot~\cite{few_shot} learning on many tasks~\cite{radford2018improving, siddiq2023exploring}, this opens up a new path to explore automated code repair using prompt engineering, where researchers develop methods to craft clear and concise prompts and use such models to obtain coherent and relevant responses.



In light of this, it is clear that there is a need for different techniques for advancing the state-of-the-art in repairing buggy code identified during the code review process. Specifically, our motivation is to explore the program repair capability of pre-trained models where the prompt will be crafted using both the buggy code and its code review.
Thus, in this paper,  we investigate the following research questions:
\begin{itemize}[leftmargin=25pt]
  \item[\bf RQ1]  \textbf{\textit{How do  pre-trained models perform in repairing bugs identified in the code review process?}}
  \item [\bf RQ2] \textbf{\textit{How effective is automated program repair using zero-shot and few-shot learning-based prompt engineering on Large Language Models?}}
  \item [\bf RQ3] \textbf{\textit{How effective are language  models in repairing bugs identified in the code review process from a developer's perspective?}}
\end{itemize}

Our work focuses on repairing buggy Java code identified during the code review process.
In the first research question, we compare two pre-trained transformer models (PLBART~\cite{PLBART} and CodeT5~\cite{CodeT5}) on the datasets of previous studies~\cite{tufano21, R4R} by fine-tuning them with the buggy codes, their fixes, and corresponding code reviews. 
For the second research question, we investigated how two LLMs (GPT-3.5-Turbo~\cite{web:openai:docs} and Code-DaVinci-Edit-001~\cite{web:openai:compatibility}) perform on these datasets using zero-shot and few-shots prompting. In the last research question, we manually investigated the output from the fine-tuned models and prompted LLMs to see how the repaired program aligned them with the code review.


The \textbf{contributions} of our work are:
\begin{itemize}[leftmargin=*]
\item Validation of the significant improvement of code repair using large language models pre-trained with NL and PL from the Tufano \etal~\cite{tufano21} dataset and the Review4Repair dataset
\item Discussion on how the architecture and pre-trained weights contribute towards the code repair performance boost.
\item Comparison of the performance of two pre-trained models, \textit{PLBART} and \textit{CodeT5}, in terms of accuracy.
\item A comprehensive investigation of two LLMs (\textit{GPT-3.5-Turbo}, and  \textit{Code-DaVinci-Edit-001}) for zero-shot and few-shot code repair with the help of prompt engineering.
\item Manual analysis of the repaired codes to understand the actual capabilities of the learning models.
\item A replication package with all the scripts used to gather the data and results\footnote{\url{https://doi.org/10.5281/zenodo.8122636}}

\end{itemize}

\section{Background}
This section explains concepts that are relevant to understand this paper.

\subsection{Code Reviews and Automated Program Repair}

\textbf{Code review}~\cite{bird2013expectations} is a software quality assurance activity in which one or more  developers analyze a peer developer's source code by viewing or reading the code parts after implementing a feature or fixing a defect. During this activity, a reviewer may identify bugs in the code. For instance, Listing \ref{lst:buggy_code_example} has a source code under review.  This example is taken from a dataset from a prior study~\cite{tufano21}, which includes the  \codeTag{\small START\normalsize} and \codeTag{\small END\normalsize} tags to indicate where a reviewer made a comment to repair a bug.
The reviewer states that the \codeJava{if} condition in line 2 ``\textit{could be simplified}''. Thus, the developer fixes the code as shown in the second snippet in Listing~\ref{lst:buggy_code_example}. 

\begin{listing}[H]
{
\begin{JavaSourceCode*}{label=\textcolor{black}{\tiny \sf \bf Code during review}}
public boolean accept(Issue issue) { 
    |\codeTag{START}| if (issueShouldNotBeReported|\tikzmark{here}|(issue, excludedLinesByRule())) { |\codeTag{END}|
        return false;
    }
    return true;
}
\end{JavaSourceCode*}
\vspace{10pt}
\begin{JavaSourceCode*}{label=\textcolor{black}{\tiny{\sf \bf Fixed code based on the review}}}
public boolean accept(Issue issue) {
    return !issueShouldNotBeReported(issue, excludedLinesByRule());
}
\end{JavaSourceCode*}
}
\caption{Example of a buggy code snippet for review.}\label{lst:buggy_code_example}
\end{listing}
\begin{tikzpicture}[remember picture]
\draw[overlay, -, line width=0.5pt, blue] ($(pic cs:here) +(-85pt,-2pt)$) -- ($(pic cs:here) +(98pt,-2pt)$);
\draw[overlay, ->, line width=1pt, blue] ( $(pic cs:here) +(0pt,-11pt)$ ) -- ($(pic cs:here) +(0pt,-3pt)$);
\node [overlay, rectangle, draw, fill=white,draw=blue, align=center,inner sep=2pt] (callout) at ($(pic cs:here) +(0pt,-16pt)$) {\tiny \it ``could be simplified''};   
\end{tikzpicture}



While code review relies on human expertise to identify and repair issues, \textbf{automated program repair}  (\textbf{APR})~\cite{apr_survey} techniques aim to automatically fix software bugs without the developer's intervention~\cite{apr_patch, apr_patch_2}. APR is also referred to as \textit{\textbf{automatic patch generation}}, \textit{\textbf{automatic bug repair}}, and \textit{\textbf{automated code repair}}. Henceforth, we will use the terms \textit{automated code repair} and \textit{automated program repair} interchangeably. 

By combining both \textit{code reviews} and \textit{APR} techniques, developers can leverage the strengths of each to enhance the overall quality of the code. 
In this work, we focus on studying how language models can automate the repair of bugs that were identified during code review.

\subsection{LLMs, Zero Shot and Few Shot Prompting}
A \textbf{{Large Language Model}} (LLM)~\cite{LLM} refers to a sophisticated artificial intelligent model which consists of a neural network with tens of millions to billions of parameters. LLMs are trained on vast amounts of unlabeled text using self-supervised learning or semi-supervised learning \cite{gpt3}.
As opposed to being trained for a single task (such as sentiment analysis or mathematical reasoning), LLMs are general-purpose models that excel in a variety of natural language processing tasks, including language translation, text generation, question-answering, summarization, and much more. GPT-3~\cite{gpt3}, BERT~\cite{devlin-etal-2019-bert}, T5~\cite{RaffelT5}, CodeBERT~\cite{feng-etal-2020-codebert} are examples of  well-known LLMs.

To direct a model's answer generation, one must carefully craft input instructions. The act of constructing and enhancing prompts to produce desired outputs is known as \textbf{{prompt engineering}}~\cite{Liu_prompting}.  
When engineering a prompt, one   may include a few input-output example pairs (\textbf{few-shot prompting}) or simply have a high-level description about the desired task (\textbf{zero-shot prompting}). A model is able to successfully complete a task due to zero-shot  and few-shot learning, which are  learning techniques that address the challenge of training models with limited training data. 




\textbf{\textit{Zero-shot learning}} is the capacity of a machine learning model to carry out a task without any explicit examples or labeled data for that specific task during training~\cite{zero_shot, zero-shot}. Large-scale code repositories specific to the target programming language are often used to train traditional code generation models. In this process, the models learn the specific syntax and semantics of that particular programming language. 
With the help of zero-shot learning, the models can be designed to generalize their understanding across various programming languages. It uses the shared concepts and patterns among many programming languages for a target language it has not seen during the training process.  


\textbf{\textit{Few-shot learning}}~\cite{few_shot} is a method where the model is trained with a limited dataset. Unlike the common practice of ML models, where the models are fed as much data as possible, few-shot learning aims to generate a model's prediction with less training data. Few-shot learning allows the model to generalize and make accurate predictions on new classes with only a few examples available for each class.


To clarify the differences between zero-shot and few-shot prompting, consider the prompt  in 
Listing \ref{lst:few_shot_prompt_example}. On one hand, lines 1--24 are a case of \textit{few-shot prompting}; it includes three examples of a buggy code, its corresponding review, and fix as well as an explicit instruction that tells the model to refactor (fix) the code based on the provided review (line 24). On the other hand, if the prompt only has the lines 13--24 (highlighted)  then it is an example of \textit{zero-shot prompting}.

\begin{listing}[H]
{
\begin{JavaSourceCode*}{highlightlines=13-24,label=\textcolor{black}{\tiny{Prompt\_example.java}}}
|\textbf{Buggy Code:} <Buggy Code 1>|
|\textbf{Review:} just return this|
|\textbf{Fixed Code:} <Fixed Code 1>|

|\textbf{Buggy Code:} <Buggy Code 2>|
|\textbf{Review:} Just return rule.|
|\textbf{Fixed Code:} <Fixed Code 2>|

|\textbf{Buggy Code:} <Buggy Code 3>|
|\textbf{Review:} Can't we just rely on @Rule?|
|\textbf{Fixed Code:} <Fixed Code 2>|

|\textbf{Buggy Code:}|
private FirewallRule findById(List < FirewallRule > collection, String id) {
    FirewallRule result = null;
    for (FirewallRule rule: collection) {
        if (rule.id().equals(id)) { 
            |\codeTag{START}| result = rule; |\codeTag{END}|
        }
    }
    return result;
}
|\textbf{Review:} Just return rule.|
|\textbf{{Refactor the Buggy Code using the Review without comments.}}|
\end{JavaSourceCode*}
}
\caption{Zero-shot and few shots prompt example.}\label{lst:few_shot_prompt_example}
\end{listing}


\section{Methodology}

\begin{figure*}[!h]
  \centering
   \includegraphics[width=\textwidth]{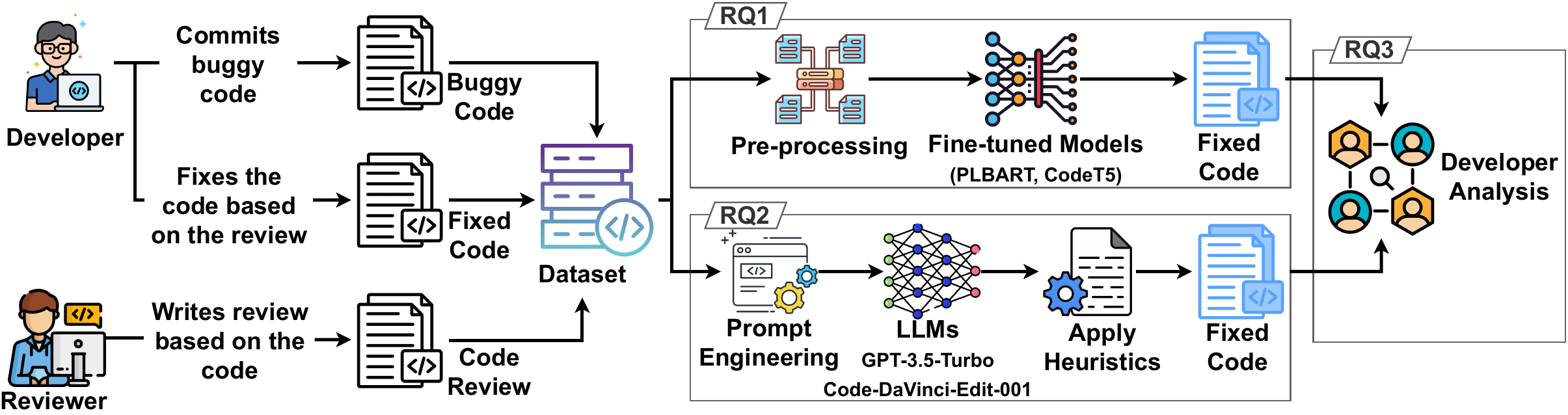}
 \caption{Overview of the Methodology.}
  \label{fig:Working_process}
\end{figure*}

Figure \ref{fig:Working_process}  provides an overview of our study. To answer \textbf{RQ1}, we collected buggy code and their code reviews from two datasets (Tufano \etal \cite{tufano21} and Review4Repair \cite{R4R}) to fine-tune two pre-trained models (PLBART and CodeT5). 
For \textbf{RQ2}, we used the same datasets in \textbf{RQ1} and prompt engineering with two LLMs (GPT-3.5-Turbo and Code-DaVinci-Edit-001). Finally, two developers conducted a manual analysis of the output of the models to check the alignment in addressing the code review in the repaired program (\textbf{RQ3}).
The next subsections explain each of these steps in detail.




\subsection{RQ1: Fine-tuning Pre-trained Models for APR}

\subsubsection{Dataset Collection and Preprocessing}

We used two datasets for repairing codes using code reviews. An overview of each dataset is given in Table \ref{tab:dataset}.  Both datasets are from recent prior works~\cite{tufano21,R4R} and consist of real examples of code reviews collected from Gerrit and GitHub. 
We preprocessed each dataset as follows:

\begin{itemize}[leftmargin=*]
    \item \textbf{Tufano \etal~\cite{tufano21} dataset}: It contains \textbf{17,194} samples of buggy code, their corresponding fixes, and code reviews collected from Gerrit and GitHub. Additionally, each buggy code has two special tokens (\codeTag{\small START} and \codeTag{\small END\normalsize}) to encapsulate the erroneous code block. 
    Similar to another study~\cite{R4R}, during the dataset preprocessing, we concatenated the buggy code and its respective code review into a single line, with the code review encapsulated using the tags \codeTag{\small |startcomment|\normalsize} and \codeTag{\small{|endcomment|}}. These concatenated snippets were the models' input, and their respective fixed codes were the target for the PLBART and CodeT5 models. We also classified the entire dataset into three fix categories: \textit{Insert}, \textit{Delete}, and \textit{Update}. These categories indicate whether the fixes only added new changes (\textit{insert}), removed code blocks (\textit{delete}), or both (\textit{update}).

    \item \textbf{Review4Repair dataset~\cite{R4R}}: It contains a total of 
    \textbf{56,211}\footnote{The paper mentioned 55,060 training samples \cite{R4R}, but the replication package contains 56,211 samples.}   
    and \textbf{2,961} samples that were used for training and testing in their study, respectively. These samples were collected from Gerrit. Since the maximum input length was not more than 512 tokens for both pre-trained models, we had to remove 57 samples from the training dataset and 6 samples from the test dataset as these samples had more than 512 tokens. 
    Hence, the initial training dataset contained 56,154 samples, and the test dataset contained 2,955 samples. 
    Since fine-tuning pre-trained models also require a validation dataset, which was not present in this dataset, we reorganized the initial training dataset to ensure that 90\% of samples are in the training dataset, 5\% of the samples are in the test dataset, and 5\% of the samples are in the validation dataset. Thus, we had \textbf{53,198} samples in the training dataset, \textbf{2,956} samples in the validation dataset, and \textbf{2,955} samples in the test dataset in our modified dataset. We also  categorized the samples into three categories (\textit{Insert}, \textit{Update}, and \textit{Delete}). To fit each sample on a single line, extra spaces and newlines were removed from each sample's code and comments in both the training dataset and test dataset samples. Similar to the Tufano \etal~\cite{tufano21} dataset, the buggy code had two special tokens, \codeTag{\small |startfocus|\normalsize} and \codeTag{\small |endfocus|\normalsize}, to encapsulate the erroneous code block. In contrast to the whole fixed code for the target in the Tufano \etal~\cite{tufano21} dataset, the target for each buggy code was merely the repair code snippet between the \codeTag{\small |startfocus|\normalsize} and \codeTag{\small |endfocus|\normalsize} special tokens. The target for the \textit{delete} class samples was an empty whitespace, which we replaced with a special token, \codeTag{|del|}.

\end{itemize}

\begin{table}[!ht]
    \centering    
    \caption{Overview of the Datasets.}
    \footnotesize
    \setlength{\tabcolsep}{5pt}
    \begin{tabular}{cccccc}
        \toprule
        \textbf{Dataset}  &
        \textbf{Type} & \textbf{Insert} & \textbf{Delete} & \textbf{Update} & \textbf{Total} \\ 
        
        \midrule
        \multirow{3}{*}{Review4Repair \cite{R4R}}&Train & 8,718 & 4,060 & 40,420 & 53,198 \\
        & Validation & 247 & 481 & 2,228 & 2,956 \\ 
        &Test & 222 & 425 & 2,308 & 2,955 \\
        \hline
         \multirow{3}{*}{Tufano \etal \cite{tufano21}}&Train & 161 & 4,385 & 9,210 & 13,756 \\ 
         & Validation & 20 & 540 & 1,159 & 1,719 \\
        &Test & 18 & 559 & 1142 & 1,719 \\                
        \bottomrule
    \end{tabular}
    \label{tab:dataset}
\end{table}

\subsubsection{Experimental Setup for Fine-tuning the Models}

After dataset preprocessing, we fine-tuned both  the \textbf{PLBART}~\cite{PLBART} and the \textbf{CodeT5}~\cite{CodeT5} models to automatically repair buggy code in a code review. 
For the PLBART model, we set the input length to \texttt{512} and the target length to \texttt{200} for both of the datasets. The settings for all other hyperparameters were identical to the standard PLBART configuration~\cite{PLBART}. 
Moreover, we used three beam sizes (\texttt{1}, \texttt{5}, and \texttt{10}) to generate the \textbf{Top-1}, \textbf{Top-5}, and \textbf{Top-10} predictions, respectively, as we experimented with various numbers of epochs to see the optimal results.
For the Review4Repair dataset~\cite{R4R}, we used \texttt{11} epochs because we found that the model's performance remains unchanged after 11 epochs. The epoch was set to \texttt{12} for the Tufano \etal~dataset~\cite{tufano21} as well because the model's performance did not increase after epoch 12. We set the hyper-parameter, \textit{patience value} to 10 epochs to observe this.
We ran these experiments in a local environment using an \textit{NVIDIA GeForce RTX 2070-8GB} GPU.

The \textit{batch size} in the CodeT5 model was set to 4, and the \textit{accumulated gradient steps} was set to 8. Furthermore, the default \textit{batch size} of 32 is ensured by the combination of batch size and accumulated gradient steps. 
Moreover, the model was fine-tuned for \texttt{45} epochs for both datasets based on  observing validation losses. We varied the hyperparameter, \textit{number\_return\_sequences} to \texttt{1}, \texttt{5}, and \texttt{10} to generate \textbf{Top-1}, \textbf{Top-5}, and \textbf{Top-10} predictions, respectively.

After tokenizing the Tufano \etal~\cite{tufano21} dataset, we observed that the maximum length of the source sequences was 590 tokens, and the maximum length of the target sequences was 194 tokens. Because the maximum input sequence length for the CodeT5 model was 512, we set the model input length to 512 and set the model output length to 200. 


Similarly, after tokenizing the Review4Repair dataset~\cite{R4R}, we observed that the maximum length of the source sequences was 561 tokens, and the maximum length of the target sequences was 116 tokens. As previously stated, we set the model input length to 512 here as well, and because the maximum target length of the sequences was 116, we set the model output length to 200. 

\subsection{RQ2: Prompt Engineering for APR}
In this section, we describe how we applied prompt engineering for both of the datasets. Next, we describe how we performed \textit{zero shot}~\cite{zero_shot} and \textit{few shot}~\cite{few_shot} prompt engineering with \textit{GPT-3.5-Turbo} and  zero-shot prompting with Code-DaVinci-Edit-001. We also detail on the heuristics used for modifying the response to fix common errors in the response from the models. 

\subsubsection{Models} \label{prompt_models}
We used two models available via OpenAI API for zero-shot prompt engineering. On the one hand,  the \textit{GPT-3.5-Turbo} is the most effective and affordable model in the GPT-3.5 family \cite{gpt3}. Although \textit{GPT-3.5-Turbo} is optimized for chat, it also performs well for code completion tasks. On the other hand, \textit{Code-DaVinci-Edit-001}~\cite{web:openai:compatibility} is another variant of Codex~\cite{chen2021Codex} GPT-3 model with editing capabilities that are specifically designed to assist with various programming-related tasks by giving instructions, including fixing code errors, completing code snippets, suggesting edits in a code snippet \etc~Given a code and an instruction in natural language, the model edits the code to comply with the instruction as close as possible.

For the few-shot prompting, we only used the \textit{GPT-3.5-Turbo} model. Since few-shots prompting gives model information about the input-output structure, it has no relation with the targeted downstream tasks. As \textit{GPT-3.5-Turbo} is a generalized model for branches of tasks, giving input-output structure with few-shots prompting can be helpful~\cite{fakhoury2023generating}. However, for the Code-DaVinci-Edit-001, there is already a fixed structure \ie input code, instruction,  and output code. Hence, there is no need for examples to make the model understand the IO structure.


\subsubsection{Zero-shot Prompt Creation}
Proper and well-crafted prompts are crucial for getting the desired response from generative models like GPT-3.5-Turbo.
For zero-shot prompt creation, we used the buggy code and the review from the respective datasets. We clearly mentioned each portion by identifying ``Buggy Code'' and ``Review'' in the prompt to make sure the model can discriminate between the buggy code and the associated review. Then we added an explicit command to fix the buggy code \textit{``Refactor the Buggy Code using the Review without comments''}. The \textit{``without comments''} clause was added to the prompt in order to ensure the response from the model does not contain any redundant or explanatory comments that were not present in the input buggy code, thus further guiding the model to produce a desired outcome. Listing \ref{lst:few_shot_prompt_example} shows the layout of the prompt for this scenario, with lines 13 through 22 designating the buggy code, line 23 designating the review associated with the buggy code, and line 24 designating the explicit command for the model to generate the fixed code respectively.

For the \textit{Code-DaVinci-Edit-001} model, we needed to pass the  buggy source code as the input of the model, and for \textit{instruction} parameter, we passed a natural language instruction \ie``\textit{Refactor the code using the Review: $<$specific code review$>$}.''

\subsubsection{Few-shot Prompt creation}
For few-shot  prompt creation, we needed to do some extra tasks. Firstly, we vectorized both the train and test dataset reviews using TF-IDF~\cite{tf-idf}. Then we calculated the cosine similarity~\cite{rahutomo2012cosine} score for each test sample with respect to every training sample. Next, we selected the three highest-ranked reviews from the training dataset and their respective buggy code, fixed code, to create the prompt for each test sample for the few-shot procedure. We aimed to feed the model three most relevant examples containing \textit{Buggy Code}, \textit{Review}, and \textit{Fixed Code} so that it could have some background knowledge while predicting the fixed code for each test sample. The later part of the prompt was the same as the zero-shot prompt. The structure of this few-shot prompt is given in Listing \ref{lst:few_shot_prompt_example}.

\subsubsection{Repaired Code Generation}
Both \textit{GPT-3.5-Turbo} and \textit{Code-DaVinci-Edit-001} models are available via the OpenAI API. 
By following recent studies~\cite{döderlein2023piloting, codex_fix_bugs, pearce_asleep_copilot, siddiq2022empirical}
for both models, we set the \textit{temperature} parameter to zero because lower temperatures cause the output to be more concentrated and deterministic. In contrast, higher temperatures cause the output to be more random.
Other parameters such as \textit{top\_p}, \textit{frequency\_penalty}, \textit{presence\_penalty} were set to their default settings, and they were 1, 0, 0, respectively.

The \textit{GPT-3.5-Turbo} model has three distinct roles: \textit{assistant}, \textit{system}, and \textit{user}.
Following the guidelines outlined in prior works~\cite{siddiq2023exploring}, we set the content for the \textit{system} role as \textit{``You are a coding assistant. You generate only the source code.''}
The content for the \textit{system} role helps the model to shape the personality of the \textit{assistant} or how it should behave for output generation. 
The prompt mentioned in the previous section was set as the content for the \textit{user} role. Finally, the \textit{assistant} role provides the fixed code as the response. For the \textit{Code-DaVinci-Edit-001}, we have the repaired code in the output directly.

\subsubsection{Heuristic-based Analysis of the Generated Repairs}\label{subsec:response_analysis}

\begin{table*}[!ht]
    \centering
        \caption{Comparison of the fine-tuned PLBART and CodeT5 models on each dataset with the respective baseline models.}
     \resizebox{2\columnwidth}{!}{%
\begin{tabular}{llccccc}
        \toprule
        \textbf{Dataset} &
        \textbf{Model Name} &
        \textbf{Top-1 Accuracy (\%)} &
        \textbf{Top-5 Accuracy (\%)} &
        \textbf{Top-10 Accuracy (\%)} &
        \textbf{BLEU-4 (\%)} &
        \textbf{CodeBLEU (\%)} \\ 
        
        \midrule
        \multirow{3}{*}{Review4Repair\cite{R4R}}
        & R4R\_CC & 19.59$_{baseline}$ & 27.73$_{baseline}$ & 31.51$_{baseline}$ & 24.66$_{baseline}$ & 39.30$_{baseline}$ \\ 
        & Fine-tuned PLBART & 25.28\positive{5.69} & 37.29\positive{9.56} & \textbf{41.42}\positive{9.91} & 40.97\positive{16.31} & 49.60\positive{10.3} \\
        & Fine-tuned CodeT5 & \textbf{29.82}\positive{10.23} & \textbf{37.73}\positive{10.0} & 39.96\positive{8.45} & \textbf{45.98}\positive{21.32} & \textbf{53.19}\positive{13.89} \\

        \cmidrule[\heavyrulewidth]{1-7}
        \multirow{3}{*}{Tufano \etal \cite{tufano21}}
        & Tufano 2-encoder    
        & 12.16$_{baseline}$ & 24.55$_{baseline}$ & 30.72$_{baseline}$ & 81.80$_{baseline}$ & 80.52$_{baseline}$ \\ 
        & Fine-tuned PLBART & 32.98\positive{20.82} & 47.12\positive{22.57} & 51.13\positive{20.41} & \textbf{87.55}\positive{5.75} & 85.46\positive{4.94} \\ 
        & Fine-tuned CodeT5 & \textbf{33.28}\positive{21.12} & \textbf{50.20}\positive{25.65} & \textbf{55.44}\positive{24.72} & 86.96\positive{4.84} & \textbf{86.80}\positive{6.28} \\
        \bottomrule
    \end{tabular}
     }
    \label{tab:result-comparison-table}
\end{table*}

We generated all the fixed codes with their respective buggy code alongside reviews with a fixed user prompt for all the datasets. We observed for the \textsl{GPT-3.5-Turbo} model that, at first, the accuracy of generating fixed code was low. However, the predicted code was somewhat similar to the target code. 
We also observed that LLMs \textit{(i)} generate repairs that had trivial syntax problems; \textit{(ii)} add an explanation of the code at the end; \textit{(iii)} generate the buggy code and fixed code together; \textit{(iv)} add a prefix \textit{java} at the first of the code; \textit{(v)} add a title before generating fixed code such as \textit{``Refactored Code''}, \textit{``Fixed Code''}~\etc~ \textit{(vi)} added extra spaces that were not needed; \textit{(vii)} enclosed the fixed code within backticks \textbf{\textasciigrave\textasciigrave\textasciigrave}. However, we could easily extract the fixed code from the response through heuristics. Hence, similar to a recent study~\cite{siddiq2023exploring}, we developed five heuristics to automatically fix the aforementioned issues:
\begin{enumerate}[leftmargin=*]
    \item[\textbf{H1}] \textbf{Adjust space}: Following the structure of the target code of Tufano \etal~\cite{tufano21} dataset, we needed to modify the response of the LLMs by removing the newlines and remove the extra spaces
    \item[\textbf{H2}] \textbf{Code explanation removal}: \textit{GPT-3.5-Turbo} sometimes explains the whole code after the fixed code generation using some keywords such as \textit{Explanation}, \textit{Reasoning}, and \textit{Changes Made}. Hence, the heuristic removes the code explanation automatically at the end alongside the keywords.
    \item[\textbf{H3}] \textbf{Remove starts with java}: \textit{GPT-3.5-Turbo} often mentions the language of the code in its response, Since our datasets only had java codes, we applied a heuristic to remove the first part of the response that starts with java.
    \item[\textbf{H4}] \textbf{Remove redundant keywords}: It removes the keywords such as \textit{Refactored code}, \textit{Corrected code}, \textit{Updated code} \etc~at the beginning of the response. Also, as we had \codeTag{\small START\normalsize} and \codeTag{\small END\normalsize} in our buggy code to specify the code block to fix, \textit{GPT-3.5-Turbo} sometimes predicts it also in the response, which was removed as they were redundant.
    \item[\textbf{H5}] \textbf{Removing backticks}: \textit{GPT-3.5-Turbo} often responds to the code snippet in markdown format where the code snippet is enclosed with backticks (\textbf{\textasciigrave\textasciigrave\textasciigrave}). Hence, we applied heuristics to remove such backticks from the response.

\end{enumerate}
We applied the above-mentioned heuristics to adjust the response to the desired behavior as much as possible. 
However, even after applying the above-mentioned heuristics, there were still some discrepancies in the responses, which needed careful human inspection to be removed. For example, the model explained the fixed code without any keyword preceding it in one scenario. Hence, we need to delete the line from the response manually. Or sometimes, the model uses unique inconsistent patterns of text like \textit{Here's the updated code} or \textit{The updated code is below}. Therefore, we manually inspected each model's response such that these erroneous patterns are  manually removed.

\subsection{RQ3: Developer Analysis  of Generated Repairs}

In the previous sections, we described how we fine-tuned PLBART and CodeT5 as well as prompted two LLMs to get the repaired code by considering the code review and evaluation based on the ground truth. However, this ground truth may not be the only possible solution, or the repaired code may not fully capture the intention in the review. For this reason, we randomly collected 314 test samples from Tufano \etal \cite{tufano21} and 340 test samples from Review4Repair \cite{R4R} datasets in order to achieve a 95\% confidence interval and 5\% error of margin. We considered five models: top-1 solution for PLBART and CodeT5, zero-shot and few-shots prompting for \textit{GPT-3.5-Turbo} after applying heuristics, and \textit{Code-DaVinci-Edit-001}. We asked two software developers to score the generated repaired code from the five models based on fulfilling the intention of the code review. They have one year of industry experience in a Fortune 500 company and significant involvement in the code review process in software development (\ie as a developer, they submit their code for code review, and they review other developers' code). They individually gave zero if the generated repaired code did not fulfill the review and gave one if the repaired code was fully aligned with the intention of the review. We then calculated Cohen's Kappa score for inter-rater agreement~\cite{mchugh2012interrater} and presented the result based on the given score. 

\subsection{Evaluation Metrics}

To evaluate a model's performance for code synthesis, there are various evaluation metrics such as BLEU (Bilingual Evaluation Understudy)~\cite{BLEU}, and CodeBLEU~\cite{CodeBLEU}, Exact Match (EM), \etc~The BLEU score denotes the quality of a machine-translated output. The CodeBLEU score utilizes the n-gram match from the BLEU score and further considers a code's important syntactic and semantic features. An exact match (EM) denotes a complete sequence-to-sequence match between the model prediction and the target code snippet.

As the n-gram match from the BLEU score emphasizes the similarity between the target and the predictions generated from the models, a naive copy can achieve higher BLEU and CodeBLEU scores with zero exact matches.
However, in a code refinement task, getting the exact correct fix is of utmost importance, as only the correct fix can ensure the successful compilation of the code. Thus, we considered the exact match between the predicted output from our model and the target code snippet as the primary evaluation metric. We further generated multiple predictions using different beam sizes and evaluated the predictions against the baseline models. 
We measure the Top-1 Accuracy as the percentage of fixes when the topmost prediction of the model exactly matches the target code snippet. Similarly, for Top-5 or Top-10 Accuracy, we measure the percentage of fixes when any of the first 5 or 10 model predictions exactly matches the target code snippet.

We used all the above-mentioned metrics to  fine-tune the models PLBART and CodeT5. We used three of them for zero-shot and few-shot prompting: BLEU, CodeBLEU, and Top-1 Accuracy.

\section{Results}

In this section, we answer our  research questions. 




\subsection{RQ1: How do  pre-trained models perform in repairing bugs identified in the code review process?}

From Table \ref{tab:result-comparison-table}, we can see that both of the fine-tuned models outperform each of the previous baseline models by a significant margin. Both baseline models~\cite{tufano21,R4R} were trained with both a buggy code and its respective code review.

On the Review4Repair dataset, the fine-tuned PLBART model achieves 9.91\%  improvement, and the fine-tuned CodeT5 model achieves 8.45\%  improvement in terms of Top-10 Accuracy over the baseline model R4R\_CC, which is the baseline model named as \textit{model\_cc} in the  Review4Repair paper~\cite{R4R}.
In terms of relative performance, the fine-tuned PLBART model achieves 5.69\%, 9.56\%, and 9.91\% higher accuracy in Top-1, Top-5, and Top-10 predictions, and on the other hand, the fine-tuned CodeT5 model achieves 10.23\%, 10.00\%, and 8.45\% higher accuracy in Top-1, Top-5, and Top-10 predictions, respectively, than the baseline model.

On the Tufano \etal~\cite{tufano21} dataset, the fine-tuned PLBART model achieves 20.41\% improvement, and the fine-tuned CodeT5 model achieves 24.72\%  improvement in terms of Top-10 Accuracy over the baseline model named Tufano 2-encoder that is the baseline model from the paper of Tufano \etal~\cite{tufano21}. The fine-tuned CodeT5 was the best-performing model; the accuracy increases ranges from 21.12\% to 25.65\%.



\begin{figure*}[!t]
\centering

\subfigure[Insert class for Review4Repair Dataset~\cite{R4R} ]{\label{fig:R4R_insert}\includegraphics[width=0.329\linewidth]{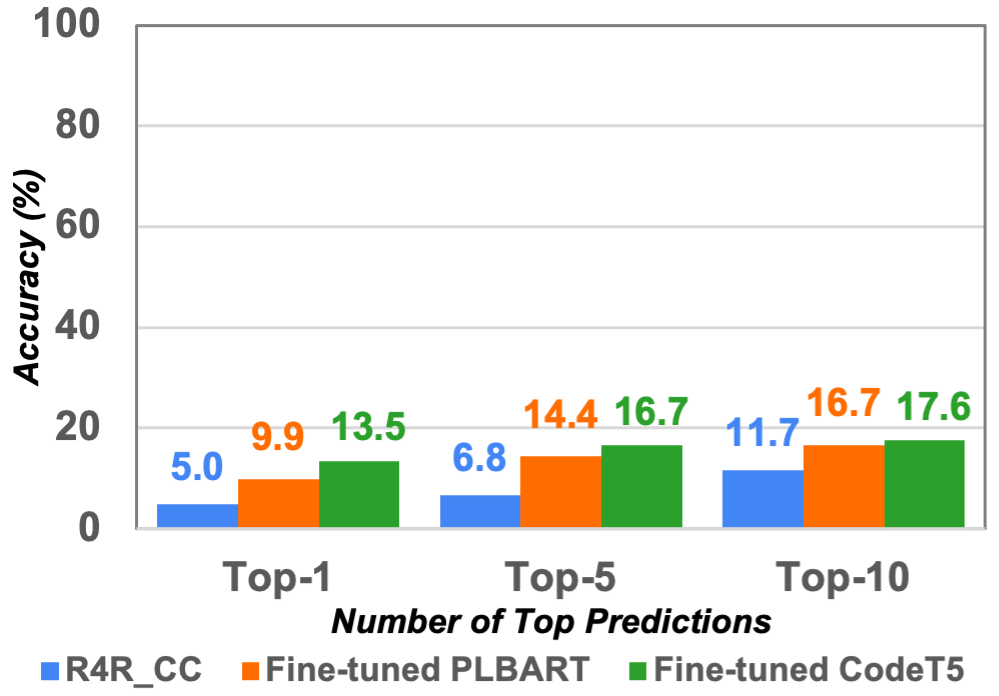}}
\subfigure[Update class for Review4Repair Dataset~\cite{R4R} ]{\label{fig:R4R_update}\includegraphics[width=0.329\linewidth]{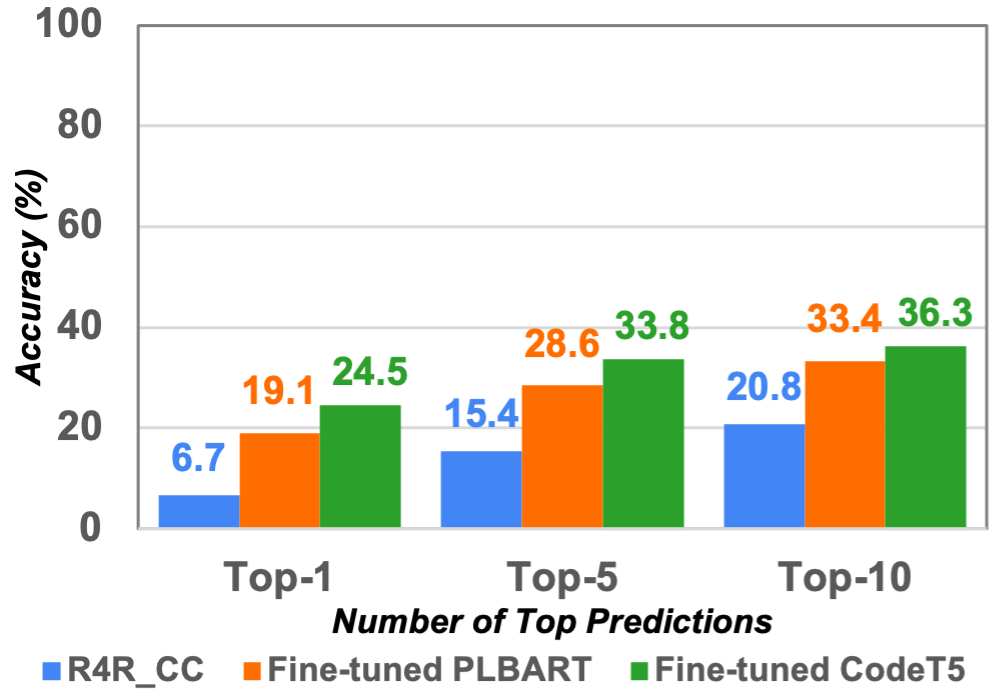}}
\subfigure[Delete class for Review4Repair Dataset~\cite{R4R} ]{\label{fig:R4R_delete}\includegraphics[width=0.329\linewidth]{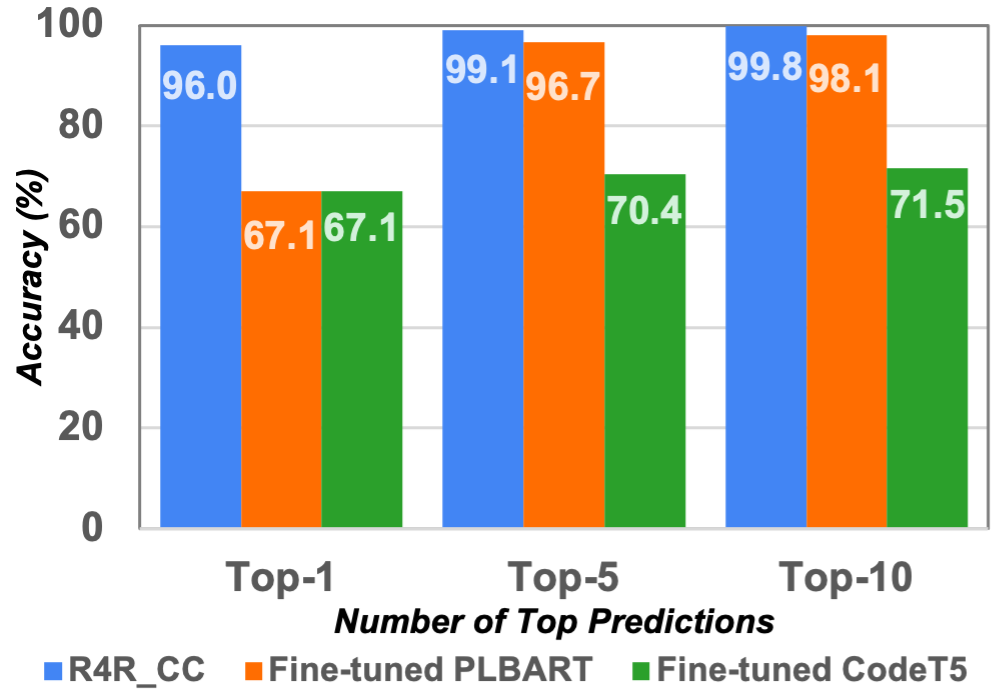}}

\subfigure[Insert class for Tufano~\etal~ Dataset~\cite{tufano21}]{\label{fig:tufano_insert}\includegraphics[width=0.329\linewidth]{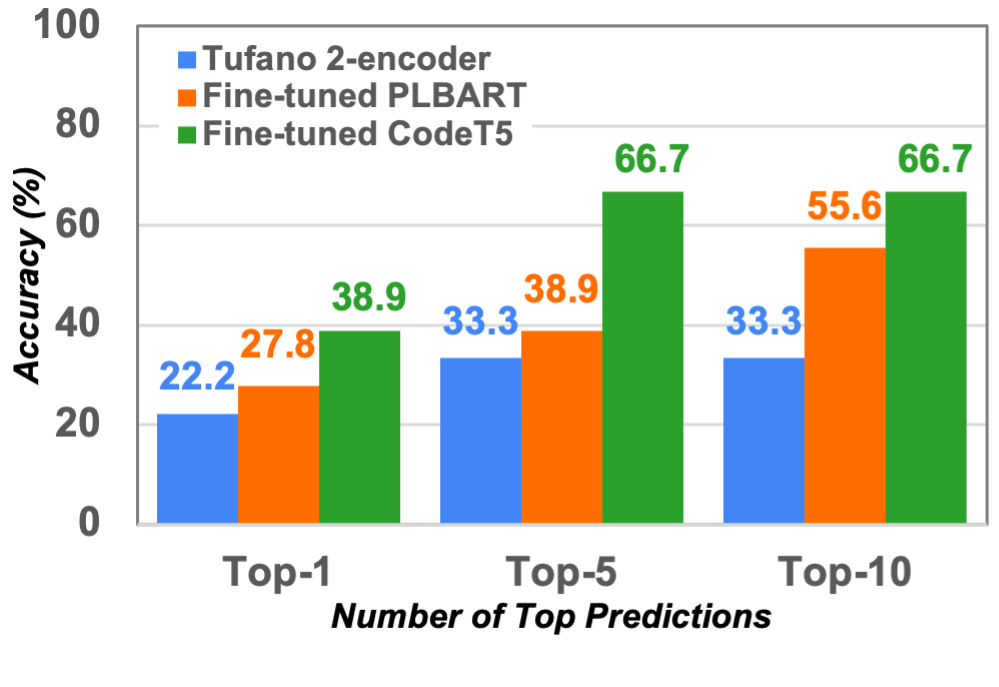}}
\subfigure[Update class for Tufano~\etal~Dataset~\cite{tufano21}]{\label{fig:tufano_update}\includegraphics[width=0.329\linewidth]{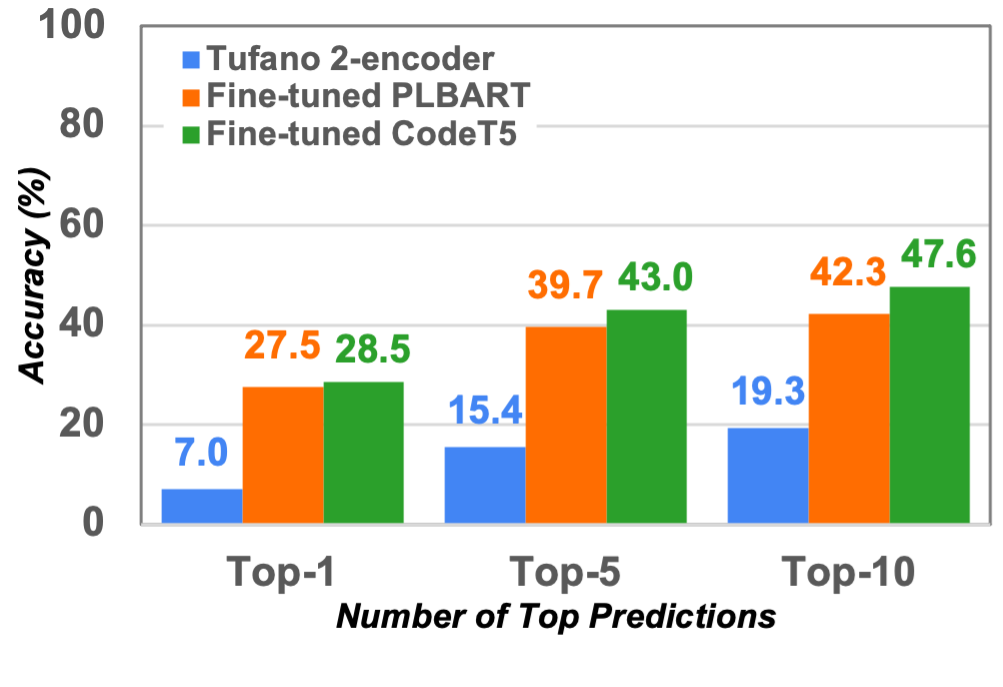}}
\subfigure[Delete class for Tufano~\etal~Dataset~\cite{tufano21}]{\label{fig:tufano_delete}\includegraphics[width=0.329\linewidth]{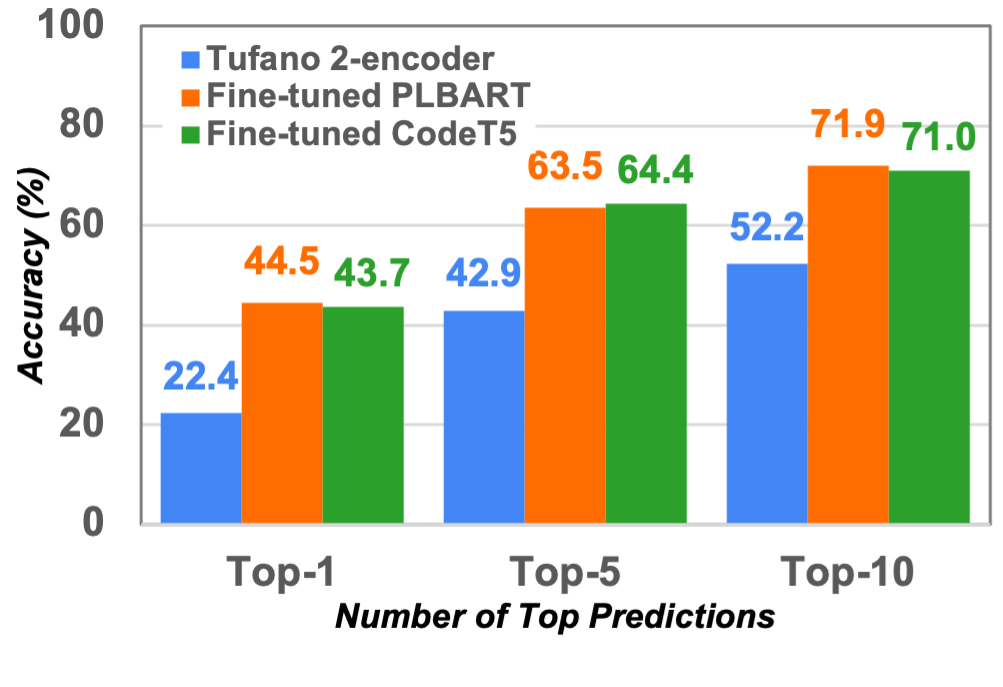}}

\caption{Performance comparison on all classes on both datasets.}
\label{fig:performance comparison on all class on both dataset}
\end{figure*}


We also report the BLEU-4 and CodeBLEU scores for each fine-tuned model in Table \ref{tab:result-comparison-table}. We can see that both the fine-tuned models improved the BLEU-4 and the CodeBLEU scores over the baseline models. This also suggests that both of the fine-tuned models can generate codes with better syntactic flow than the previous models.




%
%
%

We 

To assess each model's strengths and limitations in predicting the correct repair in all three fix categories (\ie \textit{insert}, \textit{delete} and \textit{update}), we compared the Top-1, Top-5, and Top-10 predictions generated by the fine-tuned models on both datasets for all the three classes.
Their performance is shown in Figure \ref{fig:performance comparison on all class on both dataset}.
%
%
%
From Figure \ref{fig:R4R_insert} and \ref{fig:tufano_insert}, we can see that the fine-tuned CodeT5 model achieved better accuracy in all predictions for the \textit{Insert} class than both the baseline models and the fine-tuned PLBART model for both of the datasets. This demonstrates the CodeT5 model's effectiveness in inserting additional lines of code by following the code review comments when compared to the other models.

From Figure \ref{fig:R4R_delete} and \ref{fig:tufano_delete}, on the Review4Repair dataset, the baseline model performed well in the \textit{Delete} class but poorly in the other classes. Among the two fine-tuned models, the fine-tuned PLBART model performed better than the fine-tuned CodeT5 model in the \textit{Delete} class for both datasets. This indicates that the PLBART model can perform better in removing buggy lines from code than the CodeT5 model.

From Figure \ref{fig:R4R_update} and \ref{fig:tufano_update}, we can see that in the \textit{Update} class for both of the datasets, the fine-tuned CodeT5 model outperforms both the baseline models and the fine-tuned PLBART model, similar to the performance in the \textit{Insert} Class. The \textit{Update} class requires both the insertion and deletion of specific code snippets for a correct fix. Also, for both the datasets, the \textit{Update} samples cover the larger portion. Hence, the higher performance of CodeT5 in the \textit{Update} samples leads to overall higher accuracy. Also, despite the update operation being a complicated one, the observation of the fine-tuned CodeT5 model outperforming the fine-tuned PLBART model in the \textit{Update} class suggests that the CodeT5 model can utilize the code review associated with the buggy code much better than the PLBART model. 

\begin{resultbox}
\textbf{RQ1 Findings:} Fine-tuned models can perform significantly better in generating repaired code using code review. In most cases, CodeT5 has slightly better performance than PLBART fine-tuned model. It also has comparatively better natural language and programming languages comprehension capability, and hence it can achieve better accuracy in predicting correct fixes with the help of code review than the fine-tuned PLBART model. It can be seen that predicting the correct fix for the \textit{Insert} class and the \textit{Update} class is much more difficult than for the \textit{Delete} class.
\end{resultbox}

\subsection{RQ2: How effective is automated program repair using zero-shot and few-shot learning-based prompt engineering on Large Language Models?}

We used zero-shot prompting with the code generative LLMs, GPT-3.5-Turbo and Code-DaVinci-Edit-001, and for few-shot, we utilized GPT-3.5-Turbo on both datasets.  A concise overview of these findings is presented in Table \ref{tab:gpt-result-table}

\begin{table*}[!ht]
  \centering
    \caption{Comparison of zero and few shot prompting on each dataset with the new baselines.}

  \resizebox{2\columnwidth}{!}{%
\begin{tabular}{llcccc}
    \toprule
    \textbf{Dataset} & 
    \textbf{Model Type} & 
    \textbf{Model Name} & 
    \textbf{Accuracy (\%)} & 
    \textbf{BLEU (\%)} & 
    \textbf{CodeBLEU (\%)} \\
    
    \midrule
    \multirow{7}{*}{Review4Repair\cite{R4R}}
    & \multirow{2}{*}{Pre-trained}
    & Fine-tuned PLBART & 25.28\negative{4.54} & 40.97\negative{35.41} & 49.60\negative{38.82} \\
    & & Fine-tuned CodeT5 & \textbf{29.82}\positive{0.00} & 45.98\negative{30.40} & 53.19\negative{35.23} \\
    \cmidrule(lr){2-6}
    
    & \multirow{4}{*}{Chat Style}
    & Zero-shot GPT-3.5-turbo without heuristics & 6.90\negative{22.92} & 75.42\negative{0.96} & 74.94\negative{13.48} \\ 
    & & Zero-shot GPT-3.5-turbo with heuristics & 22.06\negative{7.76} & \textbf{76.38}\positive{0.00} & 75.92\negative{12.50} \\
    & & Few-shot GPT-3.5-turbo without heuristics & 9.54\negative{20.28} & 71.55\negative{4.83} & 75.23\negative{13.19} \\
    & & Few-shot GPT-3.5-turbo with heuristics & 21.18\negative{8.64} & 71.60\negative{4.78} & 75.28\negative{13.14} \\
    \cmidrule(lr){2-6}
    
    & Instruct
    & Code-DaVinci-Edit-001 & 25.05\negative{4.77} & 75.29\negative{1.09} & \textbf{88.42}\positive{0.00} \\

    \cmidrule[\heavyrulewidth]{1-6}
    \multirow{7}{*}{Tufano \etal \cite{tufano21}}
    & \multirow{2}{*}{Pre-trained}
    & Fine-tuned PLBART & 32.98\negative{7.72} & \textbf{87.55}\positive{0.00} & 85.46\negative{3.17} \\ 
    & & Fine-tuned CodeT5 & 33.28\negative{7.42} & 86.96\negative{0.59} & 86.80\negative{1.83} \\
    \cmidrule(lr){2-6}
    
    & \multirow{4}{*}{Chat Style}
    & Zero-shot GPT-3.5-turbo without heuristics & 17.86\negative{22.84} & 70.88\negative{16.67} & 80.96\negative{7.67} \\
    & & Zero-shot GPT-3.5-turbo with heuristics & 31.70\negative{9.00} & 77.95\negative{9.60} & 83.38\negative{5.25} \\
    & & Few-shot GPT-3.5-turbo without heuristics & 27.69\negative{13.01} & 67.91\negative{19.64} & 81.03\negative{7.60} \\
    & & Few-shot GPT-3.5-turbo with heuristics & 28.21\negative{12.49} & 68.23\negative{19.32} & 81.29\negative{7.34} \\
    \cmidrule(lr){2-6}
    
    & Instruct 
    & Code-DaVinci-Edit-001 & \textbf{40.70}\positive{0.00} & 85.10\negative{2.45} & \textbf{88.63}\positive{0.00} \\
    
    \bottomrule
  \end{tabular}}
  \label{tab:gpt-result-table}
\end{table*}

We observed in zero-shot prompting that the GPT-3.5-Turbo model achieved 6.9\% and 17.86\% accuracy on the Review4Repair Dataset~\cite{R4R} and the Tufano \etal~\cite{tufano21} dataset respectively before applying the heuristics described in the Methodology section (Section \ref{subsec:response_analysis}). Comparing this performance to the fine-tuned models described in RQ1 and RQ2, it is noticeably less than ideal.

After using the heuristics, we can see a substantial improvement in accuracy.
We observed that exact match improved by 15.6\% (22.06\%-6.9\%) and 12.27\% (30.13\%-17.86\%) on the Review4Repair Dataset~\cite{R4R} and the Tufano \etal~\cite{tufano21} dataset respectively.
This implies that with proper heuristics, the model's response can be more concise and fitting for target purposes.
The improvement of BLEU and CodeBLEU score over using heuristics also implies the same.

For the case of few-shot prompting with the GPT-3.5-Turbo, this technique can provide better performance in case of accuracy and before applying heuristics. However, after applying the heuristic, this technique performs better for Tufano \etal \cite{tufano21}, but not on the Review4Repair dataset \cite{R4R}.

For the instruct model, Code-DaVinci-Edit-001, we did zero-shot prompting and it performs significantly better in some cases. For instance, it achieved state-of-the-art performance regarding the CodeBLEU score for the Review4Repair dataset \cite{R4R} and in terms of accuracy for Tufano \etal dataset \cite{tufano21}. 

\begin{resultbox}
\textbf{RQ2 Findings:} Zero-shot and few-shot prompting can be helpful when fine-tuning is not feasible. However chat-style model like the GPT-3.5-Turbo needs attention to removing an unnecessary portion in the response, whereas the instruct model like Code-DaVinci-Edit-001 has a better performance which does not need fine-tuning and heuristics to clear the output.
\end{resultbox}






\subsection{RQ3: How effective are language  models in repairing bugs identified in the code review process from a developer's perspective?}

\begin{table*}[!ht]
  \centering
    \caption{Result of the Developers' Analysis on the Repaired Code.}
    \footnotesize
\begin{tabular}{llcccc}
    \toprule
    \textbf{Dataset} & 
    \textbf{Model Type} & 
    \textbf{Model Name} & 
    \textbf{Cohen's Kappa} & 
    \textbf{Not Fulfilling} & 
    \textbf{Fulfilling} \\
    
    \midrule
    \multirow{7}{*}{Review4Repair\cite{R4R}}
    & \multirow{2}{*}{Pre-trained}
    & Fine-tuned PLBART & 0.66 &\begin{tabular}{@{}l|l@{}}
                  57.94\% & 60.29\% \\
                 \end{tabular}  & \begin{tabular}{@{}l|l@{}}
                   42.06\% & 39.71\% \\
                 \end{tabular}\\ 
    & & Fine-tuned CodeT5 & 0.68 & \begin{tabular}{@{}l|l@{}}
                   47.35\% & 51.47\% \\
                 \end{tabular}& \begin{tabular}{@{}l|l@{}}
                   52.65\% & 48.53\% \\
                 \end{tabular}\\
    \cmidrule(lr){2-6}
    
    & \multirow{2}{*}{Chat Style}
     & Zero-shot GPT-3.5-turbo & 0.51 & \begin{tabular}{@{}l|l@{}}
                   61.76\% & 60.88\% \\
                 \end{tabular}& \begin{tabular}{@{}l|l@{}}
                   38.24\% & 39.12\%\\
                   \end{tabular}\\
    & & Few-shot GPT-3.5-turbo & 0.61 & \begin{tabular}{@{}l|l@{}}
                   62.94\% & 66.18\% \\
                 \end{tabular}& \begin{tabular}{@{}l|l@{}}
                   37.06\% & 33.82\%\\
                   \end{tabular}\\
    \cmidrule(lr){2-6}
    
    & Instruct 
    & Code-DaVinci-Edit-001 & 0.61 & \begin{tabular}{@{}l|l@{}}
                   54.12\% & 58.53\% \\
                 \end{tabular}& \begin{tabular}{@{}l|l@{}}
                   45.88\% & 41.47\%\\
                   \end{tabular}\\
    \cmidrule[\heavyrulewidth]{1-6}
    \multirow{7}{*}{Tufano \etal \cite{tufano21}}
    & \multirow{2}{*}{Pre-trained}
    & Fine-tuned PLBART & 0.62 &\begin{tabular}{@{}l|l@{}}
                   45.86\% & 50.00\% \\
                 \end{tabular}  & \begin{tabular}{@{}l|l@{}}
                   54.14\% & 50.00\% \\
                 \end{tabular}\\ 
    & & Fine-tuned CodeT5 & 0.59 & \begin{tabular}{@{}l|l@{}}
                   42.99\% & 51.27\% \\
                 \end{tabular}& \begin{tabular}{@{}l|l@{}}
                   57.01\% & 48.73\% \\
                 \end{tabular}\\
    \cmidrule(lr){2-6}
    
    & \multirow{2}{*}{Chat Style}
     & Zero-shot GPT-3.5-turbo & 0.62 & \begin{tabular}{@{}l|l@{}}
                   42.36\% & 46.50\% \\
                 \end{tabular}& \begin{tabular}{@{}l|l@{}}
                   57.64\% & 53.50\%\\
                   \end{tabular}\\
    & & Few-shot GPT-3.5-turbo & 0.60 & \begin{tabular}{@{}l|l@{}}
                   45.22\% & 51.27\% \\
                 \end{tabular}& \begin{tabular}{@{}l|l@{}}
                   54.78\% & 48.73\%\\
                   \end{tabular}\\
    \cmidrule(lr){2-6}
    
    & Instruct 
    & Code-DaVinci-Edit-001 & 0.55 & \begin{tabular}{@{}l|l@{}}
                   41.08\% & 45.22\% \\
                 \end{tabular}& \begin{tabular}{@{}l|l@{}}
                   58.92\% & 54.78\%\\
                   \end{tabular}\\
    
    \bottomrule
  \end{tabular}
  \label{tab:rq4_result}
\end{table*}
To answer this research question, we have collected a statically significant amount of samples from the test of the two datasets and top results from five models to manually score them based on the fulfillment of the review in the repaired code. We presented the result in Table \ref{tab:rq4_result}. The last two columns contain the count of the score in percentages from both raters. We have 314 test samples from the Tufano \etal~\cite{tufano21} and 340 test samples from the Review4Repair dataset \cite{R4R}.
For both datasets, we can see that the raters have \textit{moderate} to \textit{substantial} agreement \cite{sun2011meta}. 

We found that, for the dataset from Tufano \etal~\cite{tufano21}, zero-shot GPT-3.5-Turbo and Code-DaVinci-Edit-001 have more capabilities in fulfilling the review in the repaired code. However, for the Review4Repair dataset \cite{R4R}, the models are comparatively less capable of addressing the reviewer's comment in the repaired code. In this case, fine-tuned CodeT5 and  Code-DaVinci-Edit-001 perform significantly better than other models.

\begin{resultbox}
\textbf{RQ3 Findings:} Language learning models face difficulties in aligning the code review in the repaired program. For Review4Repair \cite{R4R}, the fine-tuned CodeT5 can fulfill the highest 52.65\%, and for the dataset from Tufano \etal~\cite{tufano21}, Code-DaVinci-Edit-001 model can fulfill the highest 58.92\% reviews in their repaired programs.
\end{resultbox}
\section{Discussion}

In this section, we further investigate the pre-trained large language models from three different perspectives.

\subsection{Observation from the developers' analysis}
Both raters observed that the quality of reviews in the Review4Repair~\cite{R4R} dataset was not good enough to make changes as the reviews were often very vague like \textit{"nice"} or required additional context like \textit{"check my previous comment"} or \textit{"revert the change from previous commit"}
Also, in some other cases, the ground truth was not aligned with the action described in the comment, like required changes were made outside the focus scope of \codeTag{\small |startfocus|\normalsize} and \codeTag{\small|endfocus|\normalsize}.
Such scenarios could possibly lead to a difference in agreement between the two raters. 
We can notice for the Tufano~\etal~\cite{tufano21} dataset, the two developers had the most disagreement on fulfilling the CodeT5 model ($57.01\%~vs.~48.73\%$), whereas, for the Review4Repair~\cite{R4R}, they had most aligned agreement on fulfilling of the GPT-3.5-turbo model ($38.24\%~vs.~39.12\%$).
We can also notice both the developer had the highest agreement on the CodeT5 model ($\kappa=0.68$), and both the PLBART and the GPT-3.5-turbo model ($\kappa=0.62$) for the Review4Repair~\cite{R4R} dataset and the Tufano~\etal~\cite{tufano21} dataset respectively.
Also notable that both reviewers independently agreed that the CodeT5 model had the highest fulfillment for the Review4Repair~\cite{R4R} dataset and the Tufano~\etal~\cite{tufano21} dataset, the Code-DaVinci-Edit-001 achieved the highest fulfilling.

\subsection{Implication for the developers and code reviewers}
Using Large Language Models with fine-tuning and prompt engineering shows promise in the task of automating code repair. With precise and clear reviews, the models can properly interpret the intentions and be able to make the required modification.
According to our observations (\ie \textbf{RQ1}), the models struggle with more complicated code changes, such as insert and update operations, while doing significantly better for simple code changes, such as delete operations.
However, as we can see, performance improves as the number of predictions increases; thus, this can be partly addressed by having the models generate several fixes and suggestions.
Both the developer and the code reviewer can benefit from having the ability to select the most suitable fix recommendation. 

As the model's suggestions offer a starting point for making essential adjustments, this opens up ground for discussion among the developers and the code reviewers.
It is also notable that overall the performance still is not satisfactory, as shown in RQ2 and RQ3. The LLMs may make incorrect or sub-optimal suggestions. Hence, while the developers can rely on APR tools to make simpler modifications for complex code changes, both the developers and the code reviewers need to validate the model's recommendations carefully.

\section{Threats to Validity}

\textbf{Threats to internal validity} are related to how the experiments might be impacted by the model architectural settings and hyperparameter tuning. We confined our hyperparameter adjustment to modifications in batch size, source length, and target length while following the default configuration of the models for other hyperparameters. However, considering the size of the transformer architecture's search space, locating an ideal hyperparameter setting can be highly expensive. As a result, we relied heavily on the best architecture presented in both papers~\cite{PLBART,CodeT5} since the objective of our work was to fairly compare our approach's accuracy to the baseline methodologies now in use, not to determine the ideal hyperparameter configuration. We realize that there is a scope for tuning hyperparameters which is anticipated to result in more improvements.

\textbf{Threats to external validity} are related to how generalizable our results are to and across various datasets of different programming languages. We experimented and evaluated the performance of the models using the datasets from the paper of Tufano \etal~\cite{tufano21} and Review4Repair~\cite{R4R}. However, the datasets consisted of only Java codes and respective code reviews in the English language; hence, our focus was confined to a single programming language. As a result, the coverage of our findings is limited. 
Nonetheless, by using a similar methodology, other datasets of various programming languages might be investigated in future research. Additionally, we saw that GPT-3.5-Turbo, particularly the Code-DaVinci-edit model, worked remarkably well without any fine-tuning from the zero-shot or few-shot prompt engineering results. One possible reason might be that these LLMs were also trained with our aforementioned datasets. As a result, there might be a data leakage~\cite{Carlini2020ExtractingTD}. The knowledge cut-off of these two models is September 2021, where the dataset from Tufano \etal~\cite{tufano21} was published before this date, and the Review4Repair dataset \cite{R4R} was published after that. As these models are black-box, there is no way we can verify if there is data leakage for these datasets. 


\section{Related Works}
Numerous studies have been done in the past on how to automate the code repair process. 
To begin with, various studies have attempted to automate code repair without employing code review. 
Techniques like fault isolation, statement-level specification inference, and program synthesis were used in the works of \textbf{SemFix}~\cite{semfix} to generate the fixed code.
\textbf{Getafix}~\cite{getafix} employed a novel clustering algorithm to identify code changes at the AST level and utilized the context of a code change to select the most appropriate fix for a given bug. They can be used to repair SQL Injection \cite{siddiq2021sqlifix}.
In \textbf{SequenceR}~\cite{SequenceR}, copy mechanism efficiency was demonstrated and provided single-line fixes for the Java dataset.
\textbf{DeepFix}~\cite{deepfix} employed a neural network with an attention mechanism to predict fixes for common errors for programs written in the C language. 
\textbf{CoCoNut}~\cite{lutellier2020coconut} introduces the first application of the FConv~\cite{conv_seq_2_seq} architecture for automatic code repair, which removed the drawbacks of former NMT methods. Our work used different pre-trained models and LLMs to repair code based on code reviews.

Some recent works explored the importance of utilizing code review in the task of automating program repair.
\textbf{Tufano et al.}~\cite{tufano21} demonstrated this by employing two transformer models(1-encoder and 2-encoder) where the first model used only the source buggy code as input and the second model used both the source buggy code and the code review as input.
\textbf{Review4Repair}~\cite{R4R} also followed similar approach using a pointer generator network~\cite{gehrmann-etal-2018-bottom} which is a sequence-to-sequence~\cite{seq2seq} architecture for text summarization.
They also employed two models(\textit{model\_c} and \textit{model\_cc}) following similar standards like \textbf{Tufano et al.}~\cite{tufano21}. 
Both studies showed how utilizing the code review boosted the performance of their second model by a significant margin, thus establishing that learning-based models can improve their performance with the help of code review rather than using just the source code to predict proper fixes. However, in our work, we extended the study by fine-tuning models, prompting LLMs, and manually analyzing the result.

Moreover, recent development of large language models like \textbf{PLBART}~\cite{PLBART}, \textbf{CodeT5}~\cite{CodeT5}
demonstrated a strong capability of understanding both NL and PL since they are trained with many datasets.
\textbf{PLBART}~\cite{PLBART}, based on the same architecture as BART~\cite{BART}, showed promising results in a variety of downstream tasks, including code summarization, code creation, and code translation as it picks up on important program properties, including syntax, identifier naming standards, and data flow during the pre-training process mentioned in their paper. 
Also, on understanding tasks like code defect and clone detection, as well as generation tasks in a variety of directions including PL-NL, NL-PL, and PL-PL, \textbf{CodeT5}~\cite{CodeT5} performs noticeably better than previous techniques as they used two novel techniques named identifier-aware pre-training and bimodal dual generation. Our work demonstrated their usability in generating repaired code based on code review.

Furthermore, recently various large languages models like CodeGen~\cite{nijkamp2022codegen}, Codex~\cite{chen2021Codex}, and GPT-3~\cite{gpt3} showed impressive performance on code generation tasks based on NL prompts.
CodeGen~\cite{nijkamp2022codegen}, trained on a large corpus of NL and PL, proposed a conversational program synthesis approach where specifications can be provided in natural language over multiple turns and the model responses with the generated code.
GPT-3~\cite{gpt3}, a large language model developed by OpenAI, showed spectacular performance in understanding natural language and generating proper code snippets from natural language descriptions.
A fine-tuned model of GPT-3 named Codex~\cite{chen2021Codex} was the base model for Github's CoPilot.
A sub-class of GPT-3 models, GPT-3.5, included models like GPT-3.5-Turbo, the base model for OpenAI's ChatGPT.
In a recent work~\cite{siddiq2023exploring}, they demonstrated an encouraging performance of zero-shot unit test generation using the GPT-3.5-Turbo model given proper instructions as a prompt.
Our study focused on how such models can be used to automate code repair in zero-shot and few-shots learning-based prompt engineering.
\section{Conclusion}

By leveraging code review comments and the higher Programming Language (PL) and Natural Language (NL) comprehension capabilities inherited from the learned parameters, a pre-trained model can perform much better in the context of automated program repair. Furthermore, this boost in accuracy is due to mostly the learned parameters of the model rather than the architecture itself. 
Both the PLBART and the CodeT5 models effectively understood both PL and NL. Consequently, fine-tuning the models enables them to understand the specific semantics of codes and the correlations with the code reviews. Thus, both outperform the prior baseline models trained on the aforementioned datasets. In addition to that, GPT-3 \cite{gpt3} based GPT-3.5-Turbo and Code-DaVinci-Edit-001 show great promise with the prompting techniques for repairing source code based on review. However, our manual analysis demonstrated that language learning models still may not be capable of fulfilling the intention of the review in the repaired code.

\bibliographystyle{IEEEtran}
\bibliography{references}

\end{document}